\documentclass[twocolumn,10pt,cleanfoot]{formatting/asme2ej}

\newcommand{\realfield}[1]{\hbox{I \kern -.25em R}^{#1}}
\newcommand {\mb}[1]{\mathbf{#1}}
\newcommand {\bs}[1]{\boldsymbol{#1}}
\newcommand{\T}{^{\mathrm{T}}}
\newcommand{\rmd}{\textrm{d}}  
\newcommand{\myhigh}[1]{\textit{\ul{#1}}}

\usepackage[font=small, labelfont=bf, singlelinecheck=off]{caption}
\usepackage{array}
\newcolumntype{P}[1]{>{\centering\arraybackslash}p{#1}}
\newcolumntype{M}[1]{>{\centering\arraybackslash}m{#1}}
\usepackage{lipsum}
\usepackage{comment}
\usepackage{textcomp}
\usepackage{gensymb}
\usepackage{amsmath}
\usepackage{amssymb}
\usepackage{mathtools}
\usepackage{commath}
\usepackage{etoolbox}
\usepackage{caption}
\usepackage{subcaption}
\usepackage{siunitx}
\usepackage[compress]{cite}
\usepackage{cuted}
\usepackage{enumitem}
\usepackage{arydshln}
\usepackage[textsize=scriptsize, bordercolor=white,backgroundcolor=gray!30,linecolor=black,colorinlistoftodos]{todonotes} 
\usepackage[normalem]{ulem}
\usepackage{soul} 
\usepackage{xcolor}
\usepackage{framed}
\usepackage{algorithm}
\usepackage{algpseudocode}
\usepackage{setspace}
\usepackage{booktabs}
\usepackage{multirow}
\usepackage{epsfig}
\usepackage{graphicx}
\usepackage{hyperref}
\usepackage{color}
\usepackage[symbol]{footmisc}
\usepackage{float}

\graphicspath{{figures/}}
\allowdisplaybreaks
\floatstyle{plaintop}
\restylefloat{table}
\renewcommand{\citedash}{--}

\begin{document}
\title{Design, Kinematics, and Deployment of a Continuum Underwater Vehicle-Manipulator System}

\author{Justin~Sitler 
    \affiliation{
        Department of Mechanical Engineering\\
        Stevens Institute of Technology\\
        Hoboken, NJ 07030\\
        Email: jsitler1@stevens.edu\\
        Student Member,~ASME}}

\author{Long~Wang \thanks{Corresponding author.}
    \affiliation{
        Department of Mechanical Engineering\\
        Stevens Institute of Technology\\
        Hoboken, NJ 07030\\
        Email: lwang4@stevens.edu\\
        Member,~ASME}}

\maketitle

\begin{abstract}

Underwater vehicle-manipulator systems (UVMSs) are underwater robots equipped with one or more manipulators to perform intervention missions. This paper provides the mechanical, electrical, and software design of a novel UVMS equipped with a continuum manipulator, referred to as a continuum-UVMS. A kinematic model for the continuum-UVMS is derived in order to build an algorithm to resolve the robot's redundancy and generate joint space commands. Different methods to optimize the trajectory for specific tasks are proposed using both the weighted least norm solution and the gradient projection method. Kinematic simulation results are analyzed to assess the performance of the proposed algorithm. Finally, the continuum-UVMS is deployed in an experimental demonstration in which both teleoperation and autonomous control are tested for a given reference trajectory.
\end{abstract}

\section{Introduction}
\label{section:intro}

Underwater robotics is a growing field of research with practical applications including exploration \cite{trotter2019exploration}, inspection \cite{fernandes2015pipeline}, and aquaculture \cite{rundtop2016aquaculture}. Recently there has been a push in this field towards the development of intervention capabilities for free-floating underwater vehicles by integrating one or more manipulators, creating an underwater vehicle-manipulator system (UVMS). Design and control of a free-floating UVMS is a very challenging problem due to the physical design requirements imposed by the harsh environment, presence of significant and unpredictable disturbances, difficulty in modeling, and environmental barriers to communication.

An important motivation for this research is the development of low-cost solutions for underwater robotics. Many examples of autonomous underwater vehicles (AUVs) and UVMS are large and expensive, such as the Girona 500 I-AUV \cite{ribas2011gironamultipurpose, ribas2012gironasurvey}, or SAUVIM \cite{yuh1998designSAUVIM, marani2009intervention}. To lower the barrier of entry, there has been recent work towards the development of small-scale and low-cost platforms such as the Blue Robotics BlueROV2 \cite{bluerobotics} remotely operated vehicle (ROV), which has been used in work by Marais et. al. \cite{marais2022flow} as well as McConnell et. al. \cite{mcconnell2022overhead}. To help lower the barrier to entry into underwater robotics, this paper provides the design and integration of a light-duty and low-cost UVMS using the BlueROV2 as the vehicle platform.

\begin{figure}[ht]
    \centering
    \includegraphics[width=0.95\linewidth]{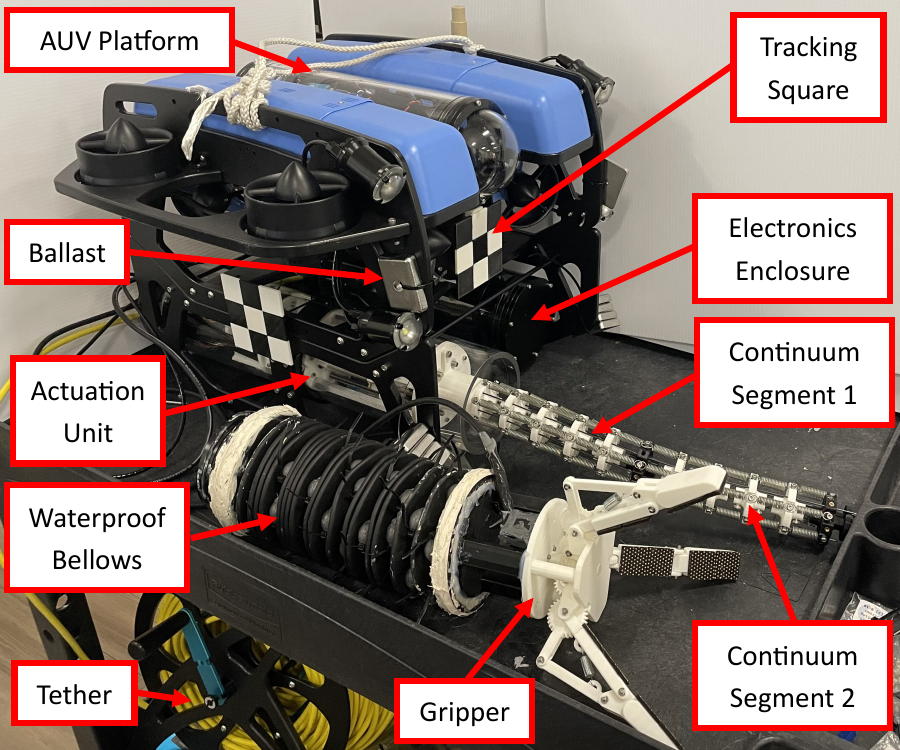}
    \caption{Continuum-UVMS integration schematic.}
    \label{fig:uvms-integration}
\end{figure}

The manipulator design proposed in this paper is a continuum manipulator, an unconventional class of robots characterized by high flexibility and compliance allowing them to achieve continuous articulating profiles. There has been limited research on continuum manipulators in field robotics or underwater applications specifically; for more detail, see Section \ref{section:related-works}. However, many features of this class of manipulators make them beneficial for underwater manipulation. The first is their natural passive compliance, which is useful in intervention tasks to compensate for positioning error and mitigate the impact of contact forces. In addition, continuum manipulators are comparatively lightweight and most of their mass is centralized in an actuation unit. This means that inertial coupling between the manipulator and the vehicle is reduced, allowing the manipulator dynamics to be neglected and thus simplifying vehicle control. Finally, the flexibility of a continuum manipulator makes them particularly useful in constrained environments such as a natural crevice or a man-made structure such as a pipe or propeller.

By combining a continuum manipulator with a free-floating vehicle platform, we introduce the novel continuum-UVMS in this paper. This design is distinguished from typical UVMSs which use serial manipulators, such as the aforementioned Girona 500 I-AUV or SAUVIM. Existing underwater continuum manipulators in the literature typically are not integrated to a free-floating vehicle \cite{lane1997AMADEUS, cianchetti2015octopus, liu2020benthic} and thus are not classified as UVMSs. Ma et. al. \cite{ma2022dualcontinuum} have proposed a similar continuum-UVMS design but have not integrated the continuum manipulator with the vehicle. Gong et. al. \cite{gong2019opposite} used a soft manipulator with an ROV platform that could be considered a continuum robot, but did not consider coordinated motions of the vehicle and manipulator. In addition, the continuum manipulator proposed in this paper is distinguished from the aforementioned design through the use of a rigid backbone and tendon actuation similar to ``elephant trunk" continuum manipulator designs \cite{hannan2001elephantdesign, hannan2003kinematics}, which have not yet been used underwater.

This paper introduces two primary contributions to UVMS design and control:

\begin{itemize}
    \item[\textbullet] Design and integration of an open-source waterproof continuum manipulator onto a free-floating underwater vehicle platform, forming a novel continuum-UVMS. In addition, the continuum-UVMS is deployed in a controlled environment to demonstrate both teleoperational and autonomous control of the UVMS.
    \item[\textbullet] Kinematics of the continuum-UVMS and development of the optimized redundancy resolution algorithm for position control of the UVMS via null-space gradient projection. The autonomous execution of the end-effector trajectory is performed both in kinematic simulation and experimentally through coordinated control of the manipulator and vehicle to highlight the impact of optimization as well as the functionality of the hardware.
\end{itemize}

This paper is outlined as follows. Section \ref{section:related-works} provides a brief literature review of the development of UVMSs, the different control and optimization methods used for these systems, and examples of continuum manipulators for various field applications. Section~\ref{section:robot-system} discusses the mechanical design of the continuum manipulator and how it is integrated with the vehicle platform. Section~\ref{section:kinematics} describes the derivation of the differential kinematics for the robotic system, implementation of the redundancy resolution algorithm, and optimization of the algorithm with respect to secondary tasks. In Section~\ref{section:simulation}, a kinematic simulation is conducted to assess the performance of the trajectory optimization. Finally, in Section~\ref{section:experiments}, the continuum manipulator and vehicle are combined in an underwater experiment to demonstrate teleoperation and autonomous control of the continuum-UVMS.

\section{Related Works}
\label{section:related-works}

\subsection{Redundancy Resolution Problem of UVMS}
UVMS missions can be broadly separated into two main categories of realistic intervention tasks \cite{ridao2015intervention}. The first task involves autonomously docking the UVMS and then performing a desired intervention task with the manipulator \cite{palomeras2014dockingintervention} while the second more challenging task involves free floating intervention \cite{ribas2015I-AUV}. These tasks have been solved by using a kinematic control layer to generate vehicle and manipulator trajectories and a dynamic control layer to achieve the desired commands. A dynamic controller was optimized for drag minimization in \cite{sarkar2001coordinated} while sliding mode control for UVMSs was implemented in \cite{yu2023adaptive}.

The kinematic control problem for a UVMS focuses on resolving the kinematic redundancy of the system, since the combined DoFs of the vehicle and manipulator exceed the minimally required DoFs of a given manipulator task in Cartesian space. This means the robot can perform additional subtasks while still satisfying its primary task. One method to do so is the gradient projection method, developed in~\cite{liegeois1977automatic} and also implemented in more recent works such as \cite{walker1988subtask, liu2010obstacle} which both used the method to avoid obstacles. Task priority control is another method to resolve this problem \cite{siciliano1991multipletasks} and has been implemented in recent works such as \cite{moe2016set}. A third method to solve the problem is through a weighted least norm solution \cite{chan1995weighted}. Other methods for subtask resolution and redundant control have been implemented as well, including combination of the weight least norm solution and gradient projection methods \cite{soylu2010redundancy} or null space saturation to avoid joint constraints and ensure feasibility of the command \cite{flacco2012motion, xing2023adaptive}.

Application of these kinematic control methods for UVMSs has been the focus of works since the early 2000's. The gradient projection method was implemented in \cite{sarkar2001coordinated,podder2004unified} to minimize drag forces. The task priority method was implemented in \cite{soylu2007dexterous, soylu2010redundancy, casalino2012agility} to maintain joint limits, manipulability, horizontal attitude, and camera field of view. A detailed summary of the modelling and control of AUVs and UVMSs can be found in \cite{antonelli2006underwater}. It is worth noting that these works are all validated in simulation; other works such as \cite{ribas2015I-AUV, simetti2014floating} validated a task priority method experimentally using the Girona 500 TRIDENT configuration.

\subsection{Manipulator Design and Continuum Manipulators}
The manipulator design is crucially important to the functionality of the UVMS. Commercially available underwater arms for use on smaller ROVs or AUVs are listed in \cite{fernandez2013grasping}; however, these designs are large and heavy rigid-link manipulators, with weights ranging from 10 kg to over 50 kg. Many of these are powered via hydraulic systems, although electric-powered arms are also available. The Graal Tech underwater modular arm (UMA) is actuated by brushless motors and has been used frequently with the Girona 500 I-AUV \cite{ribas2015I-AUV}. The cost and bulk of these manipulators have led to the development of smaller, more affordable solutions; \cite{wang2015development} designed and experimentally validated a lightweight four DoF arm and gripper actuated by servomotors contained in a centralized waterproof unit. In addition, Reach Robotics \cite{reachrobotics} has developed small, dexterous serial manipulators intended for use on small ROV platforms; for example, in \cite{marais2022flow} the BlueROV2 platform \cite{bluerobotics} was used as a small and inexpensive vehicle in conjunction with the Reach Alpha.

The continuum robot is an unconventional flexible manipulator design that can achieve continuous circular bending motion, unlike traditional manipulators which consist of a series of discrete rigid links. A classic example of a continuum manipulator is introduced in \cite{hannan2001elephantdesign} and \cite{hannan2003kinematics}. These works provide the design and kinematic control of the `elephant's trunk' manipulator, a four-segment, eight-DoF, tendon-driven continuum manipulator that mimics the motion and grasping capabilities of an elephant's trunk. The compliant and hyper-redundant design of continuum manipulators makes it difficult to estimate shape and control position, but recent works such as \cite{fang2023design, rone2018, na2020, liu2019} have addressed this problem. While much of continuum robot research has been in the lab, there are some examples of field applications of continuum manipulators as well, such as the pneumatically-actuated OctArm manipulator \cite{mcmahan2006field}.

The utility of continuum manipulators for underwater applications is discussed in \cite{davies1998subsea}, which claims their passive compliance and small number of moving parts as attractive features in an underwater environment. Potential use cases cited include grasping, visual or acoustic inspection, and bio-mimetic vehicle propulsion. A bio-inspired octopus robot with eight continuum manipulators for grasping and locomotion was proposed in \cite{cianchetti2015octopus,zheng2013}. A continuum manipulator was attached onto the SILVER2 underwater crawler for grasping small objects on the seafloor in \cite{liu2020benthic}. The AMADEUS dexterous sub-sea hand \cite{lane1997AMADEUS} uses three small hydraulically-actuated continuum fingers for an underwater gripper.  In our previous work, we developed a compact, low cost, and easy to fabricate underwater continuum manipulator \cite{sitler2022modular}, but integration with the vehicle was not addressed.

Two recent works are most relevant to our proposed continuum-UVMS design and control method. A hydraulic soft manipulator was proposed in \cite{gong2019opposite} which was attached to a free-floating underwater vehicle for delicate grasping in shallow water. It is worth noting that the vehicle was intended to be stationary on the seafloor during tasks. A prototype six DoF continuum manipulator for use on a UVMS was also introduced and tested in \cite{ma2022dualcontinuum}. The total system, which uses two arms with unique end-of-arm tooling connected to an ROV platform, is very similar to the concept and design proposed in this paper. However, it is worth noting that \cite{ma2022dualcontinuum} only provides in-air testing of the continuum manipulator.

\section{Continuum-UVMS Design}
\label{section:robot-system}

\subsection{Continuum Manipulator Design}
\label{section:mech-design}

\begin{figure*}
    \centering
    \includegraphics[width=0.85\linewidth]{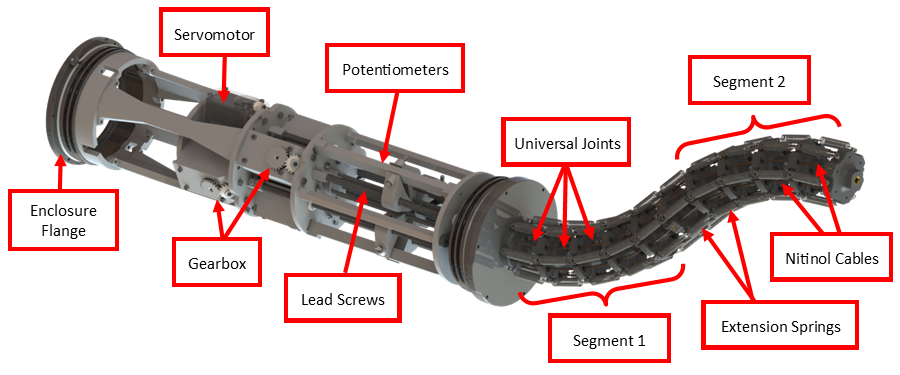}
    \caption{Two segment continuum manipulator schematic.}
    \label{fig:continuum-schematic}
\end{figure*}

\begin{figure}
    \centering
    \includegraphics[width=0.7\linewidth]{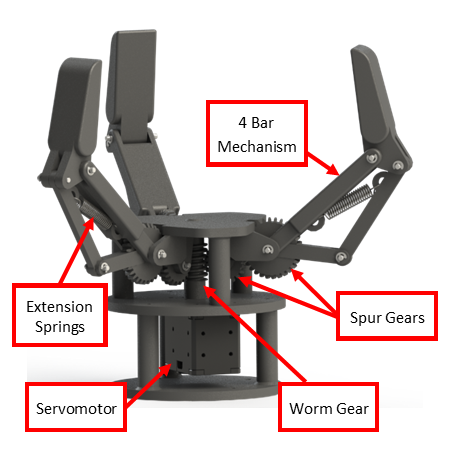}
    \caption{Underactuated gripper schematic.}
    \label{fig:gripper-schematic}
\end{figure}

The continuum robot arm used in this paper is a two-segment multi-backbone design adapted from the design developed in our previous work \cite{sitler2022modular}. The most significant change in the continuum manipulator design is the addition of a second continuum segment, giving the entire arm a total of four DoFs. This allows the distal portion of the arm to bend independently from the proximal portion, increasing its maneuverability and dexterity. The continuum manipulator is actuated via two sets of three nitinol wires which pull on the end link of each segment. The wires are connected to an actuation unit which creates the linear pulling motion via six lead screws and servo motors inside a waterproof enclosure. A gearbox was designed to change the axis of rotation and allow the servomotors to be mounted more compactly. A flexible bellows with distributed ballast covers the continuum manipulator to make the whole system waterproof and neutrally buoyant. A schematic of the manipulator can be seen in Fig. \ref{fig:continuum-schematic}, in which the enclosure tube and bellows are hidden for clarity. The CAD models, components list, and assembly instructions for this design are open-source and available in \cite{longwangwebsite}.

In addition, a 3D printed three-finger gripper is attached to the end of the continuum manipulator. This gripper, shown in Fig. \ref{fig:gripper-schematic}, uses an under-actuated design similar to the one proposed in \cite{tlegenov2014underactuatedgripper}, with a single motor actuating three 2-DoF fingers simultaneously via a 3D printed worm gear. This gear meshes with two spur gears to actuate the first DoF of the finger through a four-bar mechanism. The second DoF of each finger is actuated passively through the use of a stainless steel extension spring. The use of a worm gear allows for relatively high torque output and a simpler assembly when compared with a tendon driven under-actuated gripper. The servo motor casing is completely sealed via epoxy, and an O-ring ensures water will not leak through the motor shaft. The 3D printed and stainless steel components can be exposed to water, which means the entire gripper can be submerged without the need for a special housing. It should be noted that the printed components will absorb water over time, and prolonged exposure or use in large depth may compromise the integrity of the part; however, the proposed prototype has worked well in experiments for short periods of time in shallow water, suggesting a sufficient degree of water resistance. The motor connects to the rest of the motors and motor controller via a long waterproof cable, potted at the point of entry to ensure the arm enclosure stays sealed. While the actuation of the gripper is not considered in the kinematic control algorithm proposed in this paper, it is integrated in the experiments for verisimilitude of a real grasping task.

\subsection{Integration with Vehicle}
\label{section:communications-control}

The underwater vehicle used in this paper is the BlueROV2 by Blue Robotics, Inc. \cite{bluerobotics}. This platform was chosen because of its low cost and ease of integration via an additional payload skid, customizable waterproof enclosures and penetrators, and the Heavy Configuration upgrade, which has eight thrusters to allow for six DoF control and improved stability and maneuverability. The continuum manipulator and the electronics enclosure are mounted side-by-side on the payload skid underneath the body of the vehicle. Figure~\ref{fig:uvms-integration} labels the components of the system on the robot, with the acrylic enclosure and waterproof bellows removed for visibility.

The deployed system is broken up into four waterproof enclosures. The actuation unit contains the microprocessors, sensors, and motors required to actuate the continuum manipulator. The electronics enclosure contains a Jetson Nano computer for controlling the continuum manipulator and acts as the central hub for connecting all of the communications and power cables, as well as hosting the VectorNav VN-100 inertial measurement unit (IMU) that measures the vehicle's orientation. The BlueROV enclosure contains the flight controller, thruster controllers, and power distribution for the vehicle. This enclosure draws power from the vehicle battery stored in a separate enclosure. The BlueROV and battery enclosures are attached to the frame of the vehicle, while the actuation unit and electronics enclosure are mounted on the payload skid on the bottom of the vehicle via watertight enclosure clamps. Communication between the different enclosures is achieved via underwater cables, and communications between the entire continuum-UVMS and the topside computer is achieved via a single shielded twisted pair waterproof tether. The waterproofing enclosures and underwater cables are from Blue Robotics, Inc. \cite{bluerobotics}

The topside setup includes a Linux master computer and a tether interface. The Robot Operating System (ROS) Melodic Morenia distribution \cite{ros} is used as the software framework for establishing communications between the various components in the system. The desired end-effector pose is provided by the user and is presumed to be a stationary target pose in this paper. The kinematic control algorithm detailed in Section~\ref{section:kinematics} calculates the optimal trajectory to achieve this pose and publishes the desired vehicle state velocities, $\dot{\bs{\Psi}}_\text{V}$, and the desired continuum manipulator state, $\bs{\Psi}_\text{M}$. The vehicle control node subscribes to the desired vehicle state velocities and converts to vehicle velocity commands in the body-fixed frame $\{ \text{A}\}$, represented by $^\text{A} \dot{\bs{\Psi}}_\text{V}$. The continuum control node is adapted from our previous work \cite{sitler2022modular} and publishes motor velocity commands, $\dot{\mb{q}}$, based on the current motor position, $\bar{\mb{q}}$. Communication with the vehicle is done via ROS using the MAVLink messaging protocol on a ROS node that establishes a communications bridge. This bridge receives vehicle velocity commands and sends them to the vehicle autopilot to be executed. The code for establishing the vehicle communications and connecting to the IMU was adapted from \cite{argonautwiki} and used in works such as \cite{mcconnell2022overhead}.

\section{Kinematics for the Continuum-UVMS}
\label{section:kinematics}

For many UVMS tasks, the robot's end-effector must achieve a specific pose (position and orientation), which requires the system to have six DoFs. However, a UVMS will typically have kinematic redundancy. This means that with careful planning and decision-making, the robot can complete additional subtasks or objectives while still satisfying the primary task constraint. This section will discuss a redundant kinematic control algorithm to resolve a free-floating manipulator positioning task using a combination of the gradient projection method and weighted least norm solution. This algorithm generates a locally optimal trajectory and publishes the vehicle and manipulator commands in real time via ROS. The relevant works from which this method is adapted are discussed in Section~\ref{section:related-works}.

\subsection{Jacobian Derivation}
\label{section:jacobian-derivation}

The total Jacobian matrix relates the linear and angular velocities of the end-effector in the global coordinate system (GCS) to the vehicle and continuum manipulator state velocities. To represent the full state of the vehicle and continuum manipulator, the total Jacobian needs to account for the six DoFs from the mobile vehicle pose as well as the four DoFs from the two-segment continuum manipulator. The total state of the robot is represented by the vector $\bs{\Psi}$:
\begin{align}
    &\bs{\Psi} = 
        \begin{bmatrix}
            \bs{\Psi}_\text{V}\T & \bs{\Psi}_\text{M}\T
        \end{bmatrix}\T, \qquad 
        \bs{\Psi} \in \realfield{10\times 1}\\
    &\bs{\Psi}_\text{V} = 
        \begin{bmatrix}
            x_\text{A} & y_\text{A} & z_\text{A} & \alpha & \beta & \gamma
        \end{bmatrix}\T\\
    &\bs{\Psi}_\text{M} = 
        \begin{bmatrix}
            \theta_1 & \phi_1 & \theta_2 & \phi_2
        \end{bmatrix}\T
\end{align}

\begin{figure}[t]
    \centering
    \includegraphics[width=1.0\linewidth]{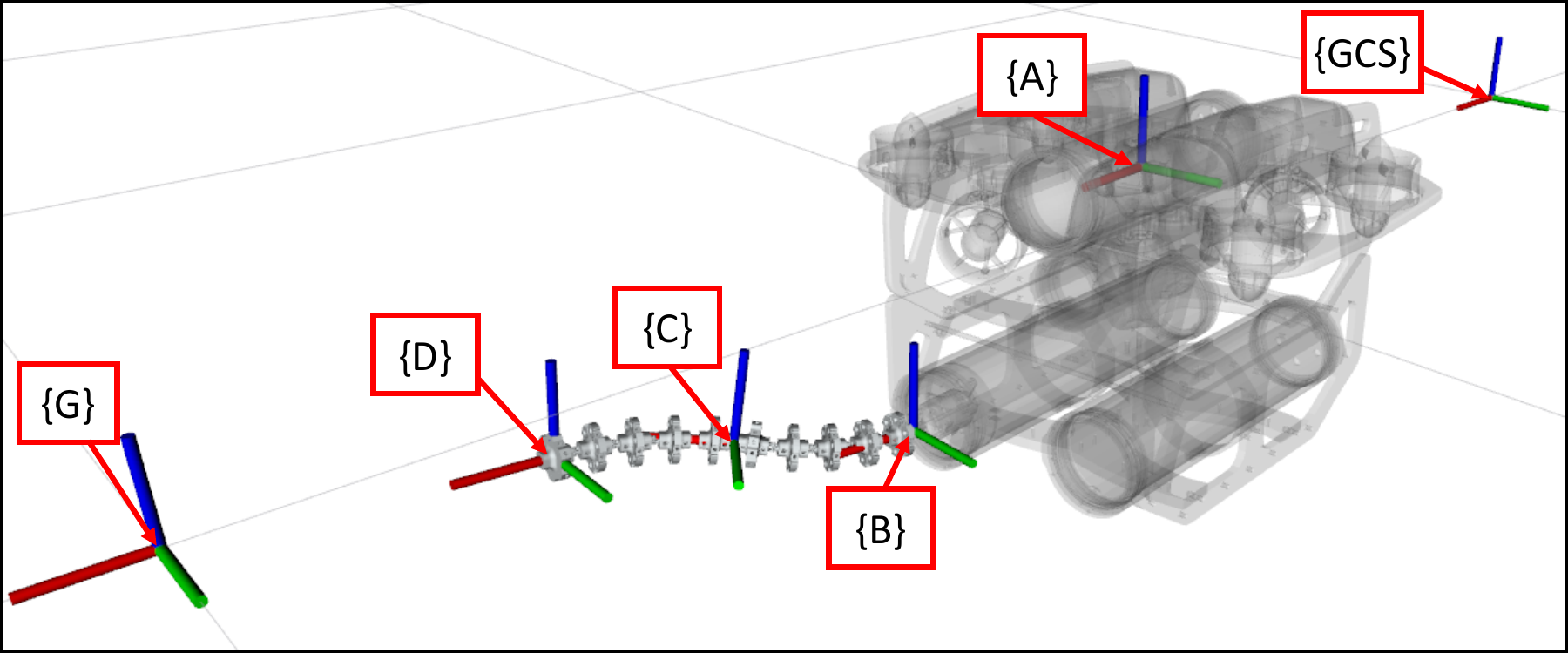}
    \caption{Continuum-UVMS coordinate frames used to define the kinematic model. \{GCS\} - global coordinate system;  \{A\} - vehicle frame;  \{B\} - base frame of $1^\text{st}$ continuum segment; \{C\} - base frame of $2^\text{nd}$ continuum segment; \{D\} - end-effector frame; \{G\} - goal frame.} \label{fig:coordinate-labels}
\end{figure}

The vector $\bs{\Psi}$ includes vehicle states $\bs{\Psi}_\text{V}$ and manipulator configuration states $\bs{\Psi}_\text{M}$. Variables $x_\text{A}$, $y_\text{A}$, and $z_\text{A}$ represent the linear position $\mb{p}_\text{A} = \begin{bmatrix} x_\text{A} & y_\text{A} & z_\text{A} \end{bmatrix}\T$ of the vehicle frame \{A\} relative to the global coordinate system (GCS), represented by frame \{GCS\}. Variables $\alpha$, $\beta$, and $\gamma$ represent respectively the yaw, pitch, and roll angles of the vehicle attitude, which can be represented in rotation matrix form by the sequence $\mb{R}_\text{A} = \mb{R}_z(\alpha) \; \mb{R}_y(\beta) \; \mb{R}_x(\gamma)$. Variables $\theta_i$ and $\phi_i$ represent respectively the bending angle and the bending plane direction angle for the $i^\text{th}$ continuum segment. The pose of the end-effector relative to frame \{GCS\} is defined as:
\begin{align}
    &\mb{x} = 
        \left\{ \mb{p}, \; \mb{R} \right\}\\
    &\mb{p} = \mb{p}_\text{A} + \mb{R}_\text{A} \; ^\text{A}\mb{r}_\text{B} + \mb{R}_\text{B} \; ^\text{B}\mb{r}_\text{C} + \mb{R}_\text{C} \; ^\text{C}\mb{r}_D
\end{align}
where $\mb{p}$ and $\mb{R}$ represent the end-effector position and orientation rotation matrix relative to frame \{GCS\}, $^\text{A}\mb{r}_\text{B}$ represents the position of the base of continuum segment 1 relative to frame \{A\}, $^\text{B}\mb{r}_\text{C}$ represents the end position of continuum segment 1 relative to frame \{B\}, and $^\text{C}\mb{r}_D$ represents the end position of continuum segment 2 relative to frame \{C\}. $\mb{R}_\text{A}$, $\mb{R}_\text{B}$, and $\mb{R}_\text{C}$ represent respectively the orientation of frame \{A\}, \{B\}, and \{C\} as shown in Figure~\ref{fig:coordinate-labels} in rotation matrix form.

We then will start deriving the total Jacobian matrix:
\begin{equation}
    \dot{\mb{x}} = \mb{J} \; \dot{\bs{\Psi}}, \quad \mb{J}\in\realfield{6\times10}\label{eqn:task-constraint}
\end{equation}
where $\dot{\mb{x}}$ is the desired end-effector twist. First, we consider the configuration-to-task space Jacobian matrix $\mb{J}_{c,i} \in \realfield{6\times2}$ for the $i^\text{th}$ continuum segment:
\begin{equation}
    \mb{J}_{c,i} = 
        \begin{bmatrix}
            \mb{J}_{p,i} \\
            \mb{J}_{\mu,i}
        \end{bmatrix}
\end{equation}
Derivation of the position and orientation Jacobians, $\mb{J}_{p,i}$ and $\mb{J}_{\mu,i}$, as well as the continuum segment displacements, $^\text{B}\mb{r}_\text{C}$ and $^\text{C}\mb{r}_D$, can be found in \cite{sitler2022modular}. These Jacobians are given by:
\begin{gather}
    \mb{J}_{p,i} = \frac{l_i}{\theta_i}
    \begin{bmatrix}
        \cos{\theta_i} - \frac{\sin{\theta_i}}{\theta_i} & 0\\
        \left(\sin{\theta_i} - \frac{\cos{\theta_i} - 1}{\theta_i}\right)\cos{\phi_i} & \left(\cos{\theta_i} - 1\right)\sin{\phi_i} \\
        \left(\sin{\theta_i} - \frac{\cos{\theta_i} - 1}{\theta_i}\right)\sin{\phi_i} & \left(1 - \cos{\theta_i}\right)\cos{\phi_i}
    \end{bmatrix}\\
    \mb{J}_{\mu,i} = 
    \begin{bmatrix}
        0 & \cos{\theta_i} - 1\\
        \sin{\phi_i} & \sin{\theta_i}\cos{\phi_i}\\
        -\cos{\phi_i} & \sin{\theta_i}\sin{\phi_i}
    \end{bmatrix}
\end{gather}
where $l_i$ represents the total length of continuum segment $i$. The transformation of these Jacobians into the GCS leads to the total Jacobian matrix:
\begin{gather}
    \mb{J} = \begin{bmatrix}
        \mb{J}_1 & \mb{J}_2 & \mb{J}_3
    \end{bmatrix}\\[6pt]
    \mb{J}_1 = 
        \underbrace{\begin{bmatrix} \mb{I}_{3 \times 3} & \left[^0\mb{r}^\text{A}_\text{D}\right]\T_{\times} \mb{T} \\
        \mb{0} & \mb{T} \end{bmatrix}}_\text{Vehicle Velocity}\\[6pt]
    \mb{J}_2 =
        \underbrace{ \begin{bmatrix} \mb{R}_\text{A} & \left[^0\mb{r}^\text{C}_\text{D}\right]_{\times} \mb{R}_\text{A} \\ \mb{0} & \mb{R}_\text{A}\end{bmatrix}\; \mb{J}_{c,1}}_{1^\text{st}\text{ Continuum Segment}}\\[6pt]
    \mb{J}_3 = 
        \underbrace{\begin{bmatrix} \mb{R}_\text{C} & \mb{0} \\ \mb{0} & \mb{R}_\text{C}\end{bmatrix} \; \mb{J}_{c,2}}_{2^\text{nd}\text{ Continuum Segment}}
\end{gather}
where $\left[\mb{v}\right]_{\times}$ represents the skew-symmetric matrix operator on vector $\mb{v}$. Matrix $\mb{T}$ represents the relation between Euler-angle rates and body-axis rates and is given by:
\begin{equation}
    \mb{T} = 
        \begin{bmatrix}
            0 & \quad -\sin{\alpha} & \quad \cos{\alpha} \; \cos{\beta} \\
            0 & \quad \cos{\alpha} & \quad \sin{\alpha} \; \cos{\beta} \\
            1 & \quad 0 & \quad -\sin{\beta}
        \end{bmatrix}
\end{equation}

This model presumes the vehicle has six degree of freedom control; however, some vehicle platforms only provide four degrees of freedom control (position and yaw). In this case, $\beta$ and $\gamma$ can be removed from $\bs{\Psi}$, and the corresponding columns (the $5^\text{th}$ and $6^\text{th}$) are removed from the total Jacobian matrix $\mb{J}$.

\subsection{Redundancy Resolution and Control}
\label{section:rr-method}

We start by considering the weighted minimum norm problem described as below:
\begin{equation}
    \label{eqn:problem-statement}
    \begin{array}{cc}
        \text{minimize } & \frac{1}{2} \; \left(\dot{\bs{\Psi}} \T \; \mb{W} \; \dot{\bs{\Psi}}\right)\\
        \text{subject to: } & \dot{\mb{x}} = \mb{J} \; \dot{\bs{\Psi}}
    \end{array}
\end{equation}
where $\dot{\mb{x}}$ is the desired end-effector twist. The choice of the weighting matrix $\mb{W}$ is discussed in more detail in Section~\ref{section:task-optimization-weight}. The weighted least norm solution to this problem is given by:
\begin{gather}
    \mb{J}_W^{+} = \mb{W}^{-1} \; \mb{J} \T \left(\mb{J} \; \mb{W}^{-1} \; \mb{J} \T \right)^{-1}\\
    \dot{\bs{\Psi}} = \mb{J}_W^{+} \; \dot{\mb{x}} \label{eqn:weighted-solution}
\end{gather}

This solution gives the state velocities that satisfy a desired end-effector twist $\dot{\mb{x}}$ while minimizing the weighted norm of the state velocity vector $\dot{\bs{\Psi}}$. The desired end-effector twist must be properly chosen in order to achieve the desired pose. We implement a resolved rates algorithm to resolve in real time the desired linear and angular velocity magnitudes $v_\text{mag}$ and $\omega_\text{mag}$ until it reaches the goal.\par
The resolved rates has been implemented alongside the redundancy resolution formulation as described in this section as well as Algorithm~\ref{alg:resolved-rates}. The desired linear velocity vector points from the current end-effector position $\mb{p}$ to the goal position $\mb{p}_\text{G} = \begin{bmatrix} x_\text{G} & y_\text{G} & z_\text{G}\end{bmatrix}\T$. As the end-effector approaches the goal, the magnitude will decrease from a constant $v_\text{max}$ when it is far away to $v_\text{min}$ as a function of the error threshold $e_p$ as well as scaling parameter $\lambda_p$. The desired angular velocity vector is applied along vector $\mb{m}$ which is found using the axis angle representation of the orientation error matrix $\mb{R}_\text{err}$, which represents the error in orientation between the current end-effector orientation $\mb{R}$ and the goal orientation $\mb{R}_\text{G}$. Like the desired linear velocity, the desired angular velocity decreases in magnitude from $\omega_\text{max}$ to $\omega_\text{min}$ as a function of the error threshold $e_\mu$ and scaling parameter $\lambda_\mu$.\par
After the end-effector twist has been calculated, the Jacobian and weighting matrices are updated to reflect the current robot state $\bs{\Psi}$. The desired robot state velocities $\dot{\bs{\Psi}}$ are calculated using the weighted inverse Jacobian and the gradient vector $\nabla g_j$, discussed in more detail in Section~\ref{section:task-optimization-gradient}. Then, the commanded robot state is updated for the next time step and published to ROS. The end-effector is considered to have reached the desired pose when the position and orientation errors are within these error thresholds, terminating the algorithm in the case of a fixed goal or continuing in the case of a moving trajectory.\par
The implementation of this algorithm is referred to in other works as the ``kinematic control layer". In this paper, we neglect the dynamics of the vehicle-manipulator system in order to focus on the kinematic modeling and motion control. The algorithm presented is differentiated from other works through the derivation of the UVMS kinematics for a continuum manipulator, and the derivation of objective functions in Section~\ref{section:task-optimization-gradient} for a continuum-UVMS.
\begin{algorithm}[t]
\caption{Redundancy Resolution with Resolved Rates}
\label{alg:resolved-rates}
\begin{algorithmic}
    \footnotesize
    \State Initialize: $\quad \mb{x}, \leftarrow \mb{x}_0$, $\quad\Psi \leftarrow \Psi_0$
    \vspace{2mm}
    \State Goal Pose: $\quad \mb{x}_\text{G} \leftarrow
        \left\{\mb{p}_\text{G}, \; \mb{R}_\text{G}\right\}$
    \State Objective Function: $\quad g(\mb{\bs{\Psi}})$
    \vspace{2mm}
    \While{$\delta_p > e_p \; || \; \delta_{\mu} > e_{\mu}$}
    \vspace{2mm}
        \State \myhigh{Step 1} - Get current pose:
        \begin{equation*}
            \mb{p}, \; \mb{R} \leftarrow \text{Kinematic Model}
        \end{equation*}
        %
        \State \myhigh{Step 2} - Compute errors:
        \begin{align*}
         &\delta_p = \sqrt{(\mb{p}_\text{G} - \mb{p}) \T \; (\mb{p}_\text{G} - \mb{p})}\\
         & \mb{R}_\text{err} = \mb{R}_\text{G} \; \mb{R}\T, 
         \quad \mu_\text{err} = \arccos{\frac{\text{Tr}(\mb{R}_\text{err})-1}{2}}\\
         &\mb{m} = \frac{1}{2 \; \sin{\mu_\text{err}}}
            \begin{bmatrix}
                \mb{R}_\text{err}^{32} - \mb{R}_\text{err}^{23} \\
                \mb{R}_\text{err}^{13} - \mb{R}_\text{err}^{31} \\
                \mb{R}_\text{err}^{21} - \mb{R}_\text{err}^{12}
            \end{bmatrix}\\
        & \delta_{\mu} = \sqrt{(\mu_\text{err} \; \mb{m}) \T \; (\mu_\text{err} \; \mb{m})}
        \end{align*}
        \State \myhigh{Step 3} - Compute desired end-effector twist:
        \vspace{2mm}
        \If{$\delta_p/e_p > \lambda_p$}
            \begin{equation*}
             v_\text{mag} = v_\text{max}   
            \end{equation*}
        \Else
            \begin{equation*}
             v_\text{mag} = v_\text{min} + \frac{(v_\text{max} - v_\text{min}) \; (\delta_p - e_p)}{e_p \; (\lambda_p - 1)}   
            \end{equation*}
        \EndIf
        \vspace{2mm}
        \If{$\delta_{\mu}/e_{\mu} > \lambda_{\mu}$}
        \begin{equation*}
            \omega_\text{mag} = \omega_\text{max}
        \end{equation*}
        \Else
            \begin{equation*}
                \omega_\text{mag} = \omega_\text{min} + \frac{(\omega_\text{max} - \omega_\text{min}) \; (\delta_\mu - e_\mu)}{e_\mu \; (\lambda_\mu - 1)}
            \end{equation*}
        \EndIf
        \begin{align*}
            & \text{Desired linear velocities:}\qquad
            \dot{\mb{p}} = v_\text{mag} \; \frac{\mb{p}_\text{G} - \mb{p}}{\delta_p} \\
            &\text{Desired angular velocities:} \qquad
            \dot{\bs{\mu}} = \omega_\text{mag} \; \mb{m} \\
            &\text{Desired twist:} \qquad \dot{\mb{x}} = 
            \begin{bmatrix}
                \dot{\mb{p}} & \quad & \dot{\bs{\mu}}
            \end{bmatrix}\T 
        \end{align*}
        \State \myhigh{Step 4} - Compute  redundancy resolution:
        \begin{align*}
            &
            \begin{matrix}
                \dot{\bs{\Psi}} =  \mb{J}_W^{+} \; \dot{\mb{x}} + \left(\mb{I} - \mb{J}_W^+ \; \mb{J} \right) \; \sum_{j=1}^{n} k_j \; \nabla g_j \left(\bs{\Psi}\right)
            \end{matrix}
            \\
            & \bs{\Psi} = \bs{\Psi} + \dot{\bs{\Psi}} \; \rmd t
        \end{align*}
    \EndWhile
\end{algorithmic}
\end{algorithm}

\subsection{Task Optimization via Weighted Least Norm}
\label{section:task-optimization-weight}

A weighted least norm (WLN) solution has been proposed in works such as \cite{chan1995weighted} to solve the redundancy problem. The diagonal weighting matrix $\mb{W}$ allows the motion of certain degrees of freedom of the robot to be prioritized or penalized. Higher value weights correspond to higher penalties on the respective degree of freedom. To start, a constant weighting matrix $\mb{W}_c$ can be used, with weights found empirically.

In this paper, two subtasks are proposed for optimization via weighting matrix: joint limit avoidance and dynamic prioritization of the continuum manipulator. The joint limit avoidance task is a classic example of optimization via the WLN method, used in works such as \cite{chan1995weighted} for a traditional robot and \cite{soylu2010redundancy} for UVMS control. For the continuum manipulator, the only limits are on the bending angles $\theta_i$, which should be less than $\theta_\text{M} = \frac{\pi}{3}$ and greater than $\theta_\text{m} = -\frac{\pi}{3}$ to prevent stalling the motors or damaging the arm. The bending plane direction angle $\phi_i$ has no limits, allowing the continuum manipulator to freely rotate in a circular path. Therefore the joint limit weighting matrix $\mb{W}_j$ should only affect the DoFs corresponding to $\theta_1$ and $\theta_2$, making these weights tend towards infinity as the manipulator approaches the joint limit:
\begin{gather}
    \mb{W}_j = \text{diag}\left(w_{j_1}, w_{j_2}, \ldots, w_{j_{10}}\right) \label{eqn:Wj}\\
    \begin{gathered}
        w_7, \; w_9 = \begin{cases}
            1 + \abs{\frac{(\theta_M - \theta_m)^2 (2\theta_i - \theta_M - \theta_m)}{4(\theta_\text{M} - \theta_i)^2(\theta_i - \theta_\text{m})^2}} & \text{if} \; \Delta \abs{\theta_i}\ge 0\\
            1 & \text{if} \; \Delta \abs{\theta_i} < 0
        \end{cases} 
        \end{gathered}
\end{gather}

For the second subtask, the free-floating robot should prioritize using the DoFs corresponding to the vehicle state $\bs{\Psi}_\text{V}$ when the end-effector is far from the target, measured by position error $\delta_p$, because the vehicle thrusters will be able to close the distance much faster than the manipulator. However, the robot should prioritize the manipulator state $\bs{\Psi}_\text{M}$ if it is close to the target, as this is more energy-efficient considering the vehicle inertia and manipulator control is generally more precise than vehicle control. A method of addressing this prioritization via WLN has been proposed in \cite{antonelli2006underwater} through the use of a weight factor $\eta$. The weight factor will be slightly modified in this paper to only consider the position error, with manipulator weights becoming large as $\eta$ approaches 1 and vehicle weights becoming large as $\eta$ approaches 0. The shape of $\eta$ as a function of $\delta_p$ should be smooth, so a fifth-order smoothing polynomial $S(x, \; x_\text{m}, \; x_\text{M})$ is used. The manipulator prioritization matrix $\mb{W}_m$ is given by:
\begin{gather}
    \mb{W}_m = \begin{bmatrix} \mb{I}_{6\times6}\left(\frac{1}{\eta}\right) & \mb{0}_{6\times4} \\ \mb{0}_{4\times6} & \mb{I}_{4\times4}\left(\frac{1}{1-\eta}\right)\end{bmatrix} \label{eqn:Wm}\\
    \eta = 0.01 + (0.9 - 0.01) \; S\left(\delta_p, 0, \lambda_\text{pre}\right) \\
    \begin{gathered}
        S(x, \; x_\text{m}, \; x_\text{M}) = 
        \begin{cases}
            1 & x \ge x_\text{M}\\
            6x^5 - 15x^4 + 10x^3 & x_\text{m} < x < x_\text{M} \\
            0 & x \le x_\text{m}
        \end{cases}
    \end{gathered} \label{eqn:smoothing-polynomial}
\end{gather}

The threshold $\lambda_\text{pre}$ will be discussed in more detail in Section~\ref{section:task-optimization-gradient}. The total weighting matrix $\mb{W}$ is found by combining the three weighting matrices:
\begin{equation}
    \mb{W} = \mb{W}_c \; \mb{W}_j \; \mb{W}_m \label{eqn:weight-total}
\end{equation}

\subsection{Task Optimization via Gradient Projection}
\label{section:task-optimization-gradient}

Works such as \cite{liegeois1977automatic} and \cite{liu2010obstacle} formulate the gradient projection method (GPM) to maximize an objective function $g_j(\bs{\Psi})$ by projecting its gradient into the null space of the Jacobian. The optimized solution for $n$ subtasks is given by:
\begin{equation}
    \dot{\bs{\Psi}} = \mb{J}_W^{+} \dot{\mb{x}} + \left(\mb{I} - \mb{J}_W^{+} \; \mb{J} \right) \; \sum_{j=1}^{n} k_j \; \nabla g_j\left(\bs{\Psi}\right) \label{eqn:optimized-solution-sum}
\end{equation}
where $\mb{J}_W^+$ denotes the weighted Moore–Penrose (pseudo) inverse of $\mb{J}$. The value of the scaling parameter $k_j$ changes how strong an affect the gradient $g_j$ has on the state velocities. In addition, a negative value of the scaling parameter results in a minimization of the objective function instead of maximization. In this paper, the scaling parameters are constant scalar values found empirically. This is a common practice for determining the scaling parameters, although other works such as \cite{walker1988subtask} and \cite{liu2010obstacle} have used a variable scaling parameter to further improve trajectory optimization. It should be noted that the weighted sum of multiple gradients does not guarantee optimal solutions for each objective function if the subtasks conflict. However, this conflict can be avoided by choosing objective functions which are orthogonal to each other.

In this paper, three subtasks will be proposed. The first is to maintain the vehicle in an upright orientation in order to easily control and maneuver it. The objective function associated with this task, $g_1$, can be defined as the dot product between the world $z$ axis, $\hat{\mb{z}}_{\text{GCS}}$, and the vehicle $z$ axis, $\hat{\mb{z}}_{\text{A}}$, which can be shown to be simply the product of the pitch and roll angles:
\begin{align}
    &g_1\left(\bs{\Psi}\right) = \hat{\mb{z}}_{\text{GCS}}\T \; \hat{\mb{z}}_{\text{A}} = \cos{\beta} \cos{\gamma}\\
    &\nabla g_1\left(\bs{\Psi}\right) = \begin{bmatrix} \mb{0}_{4\times1} \\ -\sin{\beta} \cos{\gamma} \\ -\cos{\beta} \sin{\gamma} \\ \mb{0}_{4\times1}\end{bmatrix}
\end{align}

The second subtask is to orient the vehicle towards the target. This is important in many applications to allow the camera to clearly see the target. In this paper, this task will only use the orientation in the global XY plane in order to avoid conflict with the first subtask. This means the objective function associated with this subtask, $g_2$, will be a function only of $\alpha$, $\mb{p}_\text{A}$, and $\mb{p}_\text{G}$, ensuring orthogonality with the first subtask. The objective function can be defined as the square of the difference between the vehicle yaw angle $\alpha$ and a misalignment angle $\zeta$:
\begin{gather}
    g_2\left(\bs{\Psi}\right) = \left( \alpha - \zeta \right)^2 \\
    \zeta = \arctan\left(\frac{y_\text{G} - y_\text{A}}{x_\text{G} - x_\text{A}}\right) = \arctan\left(\frac{\delta_y}{\delta_x}\right)\\
    \nabla g_2\left(\bs{\Psi}\right) = 2 \left(\alpha - \zeta\right) \begin{bmatrix}\frac{-\delta_y}{\delta_x^2 + \delta_y^2} & \frac{\delta_x}{\delta_x^2 + \delta_y} & 0 & 1 & \mb{0}_{1 \times 6}\end{bmatrix}\T
\end{gather}

The third subtask is to maintain a desired manipulator pose during the travel and grasp preparation phase of executing the trajectory. The motivation for this subtask is to keep the arm in a pose that will be easy to hold during the faster travel phase, then switch to a high dexterity pose in preparation for completing a grasp. For a continuum manipulator, singularities occur only when $\theta_i$ equal zeros; the values of $\phi_i$ do not affect the total dexterity, so we will only consider the manipulator bending configuration $\bs{\psi} = \begin{bmatrix} \theta_1 & \theta_2 \end{bmatrix}\T$ for this subtask. The objective function $g_3$ can be defined as the squared norm of the manipulator pose error:
\begin{gather}
    g_3\left(\bs{\Psi}\right) = \| \bs{\psi} - \bs{\psi}_\text{des}\|^2 = \left(\bs{\psi} - \bs{\psi}_\text{des}\right)\T\left(\bs{\psi} - \bs{\psi}_\text{des}\right)\\
    \nabla g_3\left(\bs{\Psi}\right) = \begin{bmatrix} \mb{0}_{2\times6} & \begin{array}{c} 2 \\ 0 \end{array} & \mb{0}_{2 \times 1} & \begin{array}{c} 0 \\ 2 \end{array} & \mb{0}_{2 \times 1}\end{bmatrix} \T\left(\bs{\psi} - \bs{\psi}_\text{des}\right)
\end{gather}

The definition of the desired pose $\bs{\psi}_\text{des}$ is challenging because it changes as the end-effector approaches the target. To this end, a range threshold and desired bending configuration for the travel phase, $\lambda_\text{tra}$ and $\bs{\psi}_\text{tra}$, as well as a range threshold and desired bending configuration for the preparation phase, $\lambda_\text{pre}$ and $\bs{\psi}_\text{pre}$, are defined. In order to have a smooth transition between these poses, the fifth order smoothing polynomial $S(x, \; x_\text{m}, \; x_\text{M})$ defined in Equation~\ref{eqn:smoothing-polynomial} is added:
\begin{align}
    &\bs{\psi}_\text{des} = \bs{\psi}_\text{pre} + S \left(\frac{\delta_\text{pre} - \lambda_\text{pre}}{\lambda_\text{tra} - \lambda_\text{pre}}, \; \lambda_\text{pre}, \; \lambda_\text{tra}\right)\; \left(\bs{\psi}_\text{tra} - \bs{\psi}_\text{pre}\right) 
\end{align}

Finally, the scaling parameters $k_j$ are found empirically to achieve smooth and accurate performance in simulation. Objective function $g_1$ is to be maximized while $g_2$ and $g_3$ are to be minimized, so $k_1$ should be positive while $k_2$ and $k_3$ should be negative. In addition, $k_3$ should only apply throughout the travel and preparation phase of the trajectory, and be disabled during the final approach. This can be achieved by updating the value of the scaling parameter based on the position error.

It is worth noting that these subtask objective functions are, by inspection, orthogonal in the robot's configuration space, ensuring no conflict between subtasks. In cases where choosing orthogonal objective functions is not possible, a different optimization method such as the task priority method should be used, as discussed in Section~\ref{section:related-works}.

\section{Simulation Validation}
\label{section:simulation}

\begin{figure}[t]
    \centering
    \includegraphics[width=0.95\linewidth]{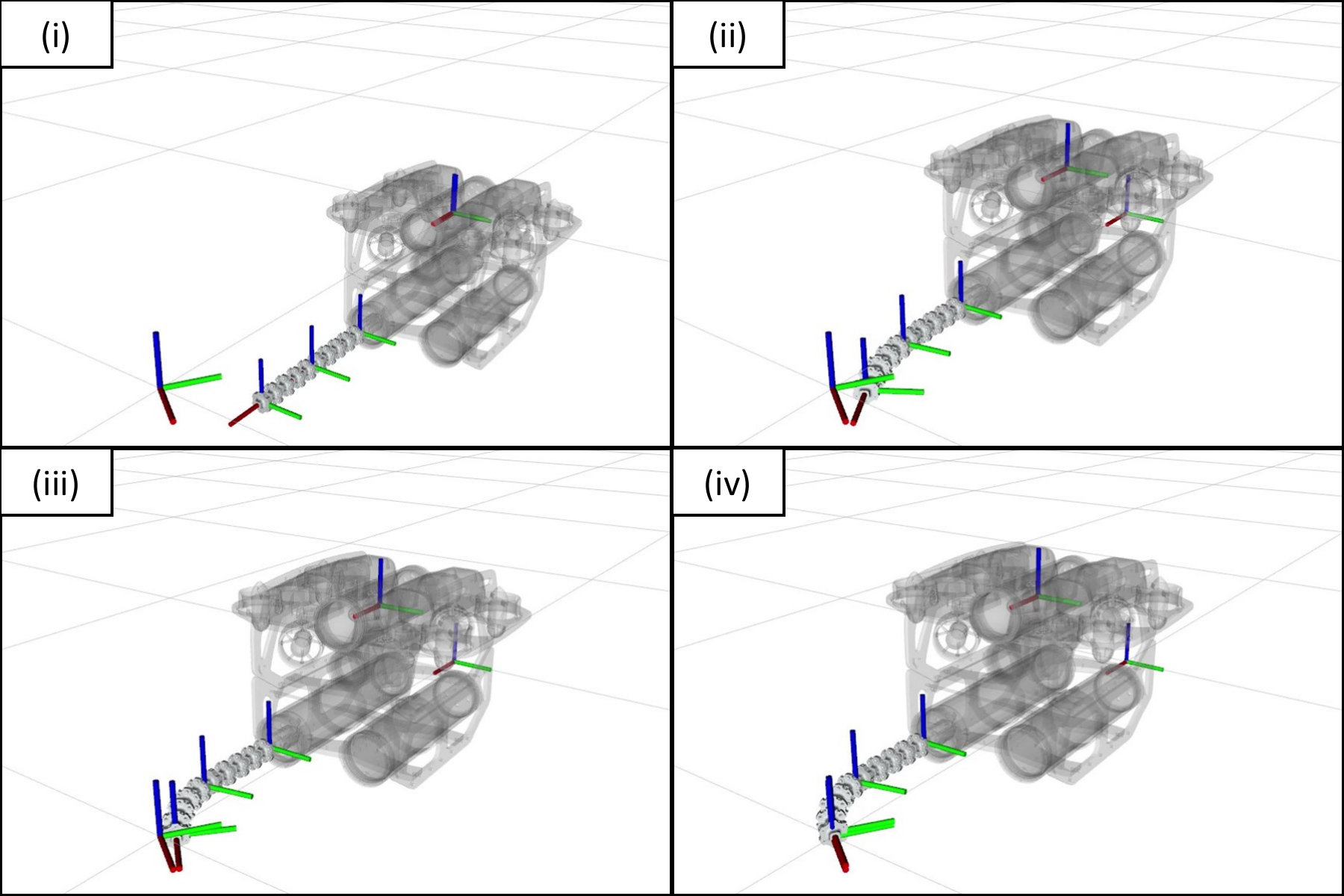}
    \caption{Time-lapse of the optimized simulation trajectory, visualized using RViz. The UVMS and goal pose coordinate frames can be seen in each frame. The starting configuration is shown in (i), and the coordinated motion of both the vehicle and manipulator can be seen between each frame until the end effector reaches the goal pose in (iv).}
    \label{fig:simulation-timelapse}
\end{figure}

To validate the algorithm proposed to resolve the redundancy of the free-floating vehicle system, two sets of kinematic simulations with multiple cases were performed using ROS Melodic Morenia and Python with a time step of 0.01 seconds. The simulation was run on Ubuntu 18.04 with Intel Core i9-9900K @ \SI{3.60}{\GHz} and 16 GB DDR4 RAM. The resolved motion rate algorithm is purely kinematic, and the simulation does not consider any dynamics since this is outside the scope of this paper. The first simulation analyzes the impact of the weighting matrix, and the second analyzes the impact of the gradient projection. The desired task for both simulations is simply to move the end-effector to a desired position of $\mb{p}_\text{G} = \begin{bmatrix} 1 & 0 & 0\end{bmatrix} \text{m}$ and orientation of $\mb{R}_\text{G} = \mb{R}_z\left(1 \;\text{rad}\right)$, imitating a basic grasp of an object at a known location. The $z$ component of the goal orientation will require the robot to bend the end-effector in order to achieve this pose while still maintaining the field of view associated with the second subtask. A snapshot showing the continuum-UVMS, goal pose, and GCS is shown in Figure~\ref{fig:coordinate-labels}. A summary of the weighting matrix and gradient scaling parameters used for each case is contained in Table~\ref{table:simulation-parameters}.

The weighting matrix comparison simulation was repeated for four cases in total. In Case 1, an identity matrix is used to equally weight all variables as a baseline for comparison. Case 2 shows the impact of manipulator prioritization using the constant weighting matrix $\mb{W}_c$. Case 3 introduces the joint limit weighting matrix $\mb{W}_j$ as defined in Equation~\ref{eqn:Wj}. Finally, Case 4 addresses dynamic manipulator prioritization by including the weighting matrix $\mb{W}_m$ defined in Equation~\ref{eqn:Wm}. All other simulation parameters are identical between cases, and all subtask objective functions are disabled by setting the scaling parameters $k_j$ to zero.

\begin{figure}[!ht]
    \centering
    \captionsetup{justification=centering}
    \begin{subfigure}[b]{0.49\linewidth}
        \includegraphics[width=\textwidth]{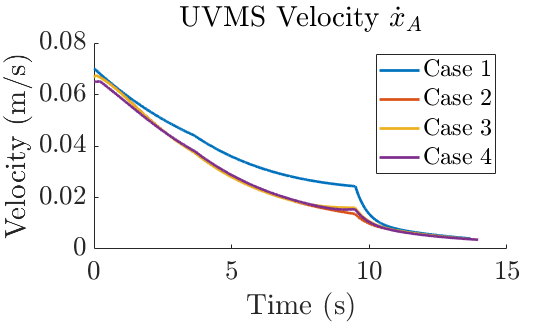}
        \label{fig:simulation-weighting-UVMS-a}
    \end{subfigure}
    \begin{subfigure}[b]{0.49\linewidth}
        \includegraphics[width=\textwidth]{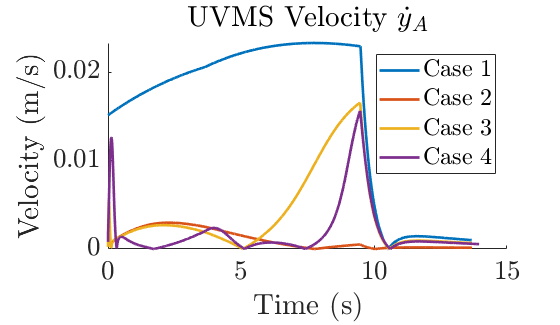}
        \label{fig:simulation-weighting-UVMS-b}
    \end{subfigure}
    \begin{subfigure}[b]{0.49\linewidth}
        \includegraphics[width=\textwidth]{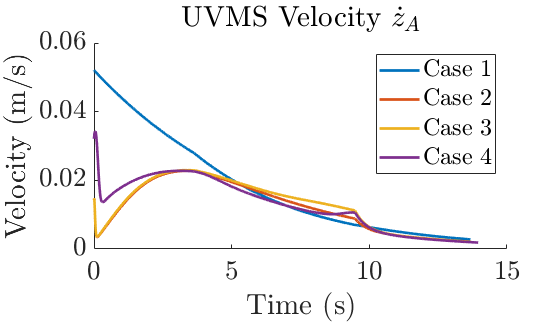}
        \label{fig:simulation-weighting-UVMS-c}
    \end{subfigure}
    \begin{subfigure}[b]{0.49\linewidth}
        \includegraphics[width=\textwidth]{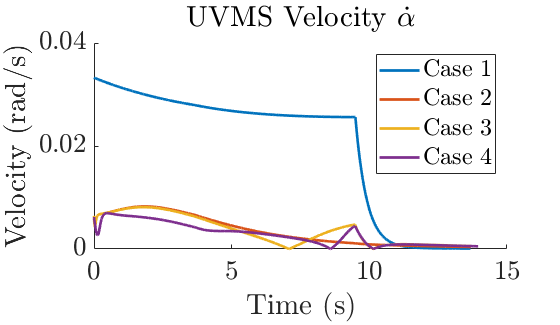}
        \label{fig:simulation-weighting-UVMS-d}
    \end{subfigure}
    \caption{Weighting matrix impact on UVMS velocity.}
    \label{fig:simulation-weighting-UVMS}
\end{figure}

\begin{figure}[!ht]
    \centering
    \begin{subfigure}[b]{0.49\linewidth}
        \includegraphics[width=\textwidth]{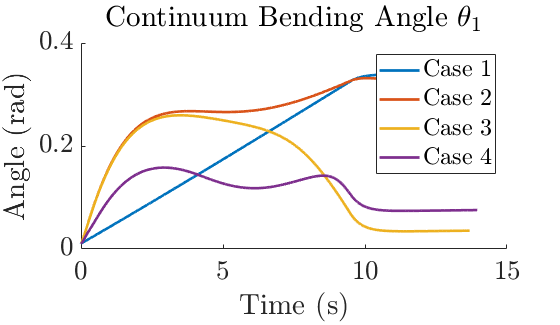}
        \label{fig:simulation-weighting-continuum-limits-a}
    \end{subfigure}
    \begin{subfigure}[b]{0.49\linewidth}
        \includegraphics[width=\textwidth]{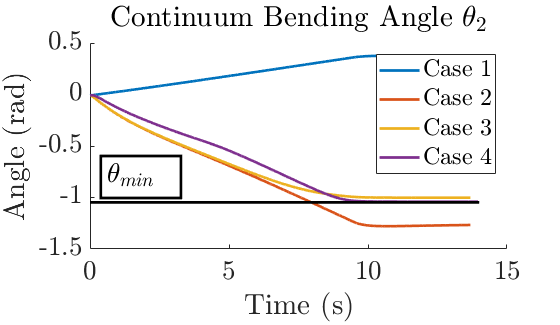}
        \label{fig:simulation-weighting-continuum-limits-b}
    \end{subfigure}
    \begin{subfigure}[b]{0.49\linewidth}
        \includegraphics[width=\textwidth]{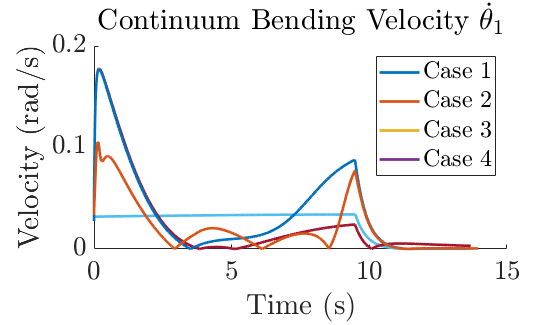}
        \label{fig:simulation-weighting-continuum-velocity-a}
    \end{subfigure}
    \begin{subfigure}[b]{0.49\linewidth}
        \includegraphics[width=\textwidth]{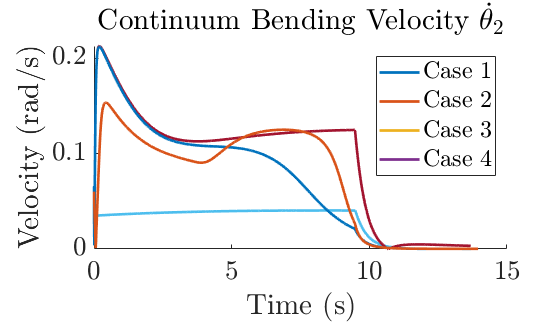}
        \label{fig:simulation-weighting-continuum-velocity-b}
    \end{subfigure}
    \caption{Weighting matrix impacts on continuum position and velocity.}
    \label{fig:simulation-weighting-continuum}
\end{figure}

The impact of each weighting matrix choice is demonstrated in Figures~\ref{fig:simulation-weighting-UVMS} and~\ref{fig:simulation-weighting-continuum}. As expected, the baseline case has the largest vehicle velocity commands while hardly utilizing the manipulator, as seen by the relatively low velocity commands. This is contrasted by much larger manipulator velocity and generally reduced vehicle velocities in Cases 2, 3, and 4. The introduction of the joint limit weighting matrix is best seen in the response of $\theta_2$ in Cases 3 and 4, where the bending angle approaches but does not reach the minimum joint limit represented by a black horizontal line. The impact of the dynamic manipulator prioritization in Case 4 is subtle, but slightly higher linear vehicle velocities towards the beginning of the simulation and slightly higher manipulator bending velocities $\dot{\theta}_i$ towards the end can be seen. The impact of the weighting matrix choice is quantitatively compared in Table \ref{table:wln-velocity-average}, which lists the average value of the vehicle velocities ($\overline{\dot{x}}_\text{A}$, $\overline{\dot{y}}_\text{A}$, $\overline{\dot{z}}_\text{A}$, $\overline{\dot{\alpha}}$) and manipulator bending velocities ($\overline{\dot{\theta}}_1$, $\overline{\dot{\theta}}_2$). Implementing the matrix $\mb{W}_c$ shows a reduction in the average vehicle velocities between Cases 1 and 2 and increase in the magnitude of the manipulator bending velocity, particularly $\overline{\dot{\theta}}_2$. Implementing $\mb{W}_j$ shows a reduction in $\overline{\dot{\theta}}_2$ between Cases 2 and 3 to avoid the joint limit on $\theta_2$. Finally, there is a small increase in magnitude of $\overline{\dot{\theta}}_1$ and $\overline{\dot{\theta}}_2$ between Cases 3 and 4 due to the introduction of $\mb{W}_m$ to further prioritize the manipulator as the robot reaches the goal pose.

The impact of the GPM to optimize the proposed objective functions is the focus of the second simulation. To demonstrate this, the simulation was repeated for five cases. In Case 5, all of the objective functions were disabled by setting their respective scaling parameters $k_j$ to zero in order to show the baseline solution without gradient projection. Next, Cases 6, 7, and 8 were run with only $g_1$, only $g_2$, and only $g_3$ enabled respectively to demonstrate the optimization of a single objective function at a time. In Case 9, all three objective functions were enabled to highlight the ability of the algorithm to handle multiple subtasks simultaneously.  The total weighting matrix $\mb{W}$ is included in each case. A time-lapse of the optimized trajectory solution to Equation~\ref{eqn:optimized-solution-sum} is shown in Figure~\ref{fig:simulation-timelapse}.

\begin{figure}[!ht]
    \centering
    \begin{subfigure}[b]{0.49\linewidth}
        \includegraphics[width=\textwidth]{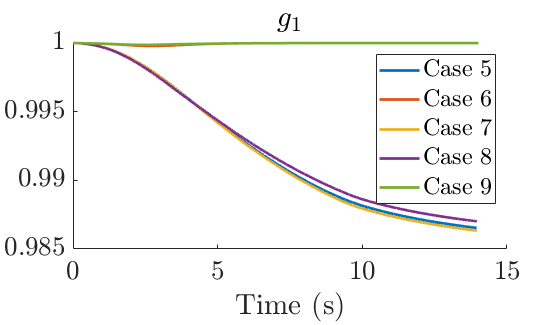}
    \end{subfigure}
    \begin{subfigure}[b]{0.49\linewidth}
        \includegraphics[width=\textwidth]{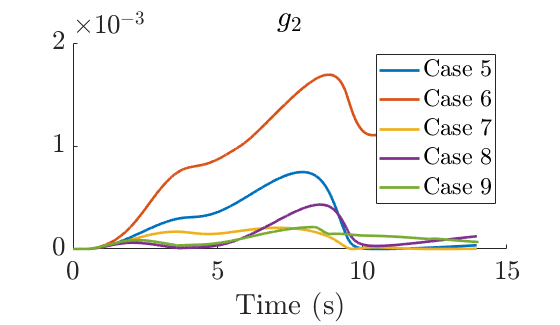}
    \end{subfigure}
    \begin{subfigure}[b]{0.49\linewidth}
        \includegraphics[width=\textwidth]{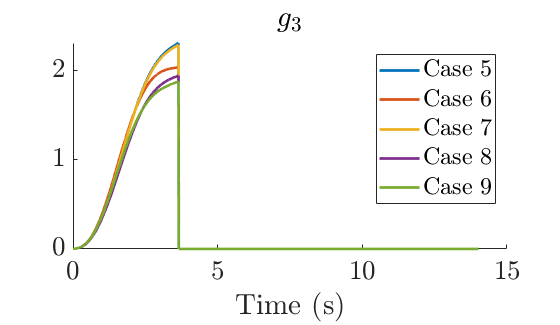}
    \end{subfigure}
    \caption{Gradient projection impacts on optimizing objective functions.}
    \label{fig:simulation-gpm-objective}
\end{figure}

The optimization of each cost function via the GPM is shown for each case in Figure~\ref{fig:simulation-gpm-objective}. The baseline solution in Case 5 is, as expected, poorly optimized for all three subtasks, with small values of $g_1$ and large values of $g_2$ and $g_3$. This trend is also evident in cases where the respective subtask is disabled; for example, Case 6 shows poor optimization of $g_2$ and $g_3$. However, this case shows a significant improvement in optimization of $g_1$. The impact of the gradient projection method is quantitatively compared in Table \ref{table:gpm-cost-average}, which lists the average values of the cost functions under each case. Compared to the baseline Case 5, it can be clearly seen that Case 6 best maximizes $g_1$, Case 7 best minimizes $g_2$, and Case 8 best minimizes $g_3$. In addition, combining all of the subtasks in Case 9 produces similar or better optimization of a given subtask than the corresponding single-subtask case.

\begin{table}[ht]
    \centering
    \footnotesize
    \begin{tabular}{ c c c c c c }
        \toprule
        & Case & Weighting Matrix & $k_1$ & $k_2$ & $k_3$ \\
        \midrule
        \parbox[t]{2mm}{\multirow{3}{*}{\rotatebox[origin=c]{90}{WLN}}} & 1 & $\mb{I}$ & 0 & 0 & 0 \\
        & 2 & $\mb{W}_c$ & 0 & 0 & 0 \\
        & 3 & $\mb{W}_c \mb{W}_j$ & 0 & 0 & 0 \\
        & 4 & $\mb{W}_c \mb{W}_j \mb{W}_m$ & 0 & 0 & 0 \\
        \midrule
        \parbox[t]{2mm}{\multirow{5}{*}{\rotatebox[origin=c]{90}{GPM}}} & 5 & $\mb{W}_c \mb{W}_j \mb{W}_m$ & 0 & 0 & 0 \\
        & 6 & $\mb{W}_c \mb{W}_j \mb{W}_m$ & 3 & 0 & 0 \\
        & 7 & $\mb{W}_c \mb{W}_j \mb{W}_m$ & 0 & -0.05 & 0 \\
        & 8 & $\mb{W}_c \mb{W}_j \mb{W}_m$ & 0 & 0 & -0.1 \\
        & 9 & $\mb{W}_c \mb{W}_j \mb{W}_m$ & 3 & -0.05 & -0.1 \\
        \bottomrule
    \end{tabular}
    \caption{Simulation parameters for weighted least norm (WLN) and gradient projection method (GPM) on each case.}
    \label{table:simulation-parameters}
\end{table}

\begin{table}[ht]
    \centering
    \footnotesize
    \begin{tabular}{ c c c c c c c }
        \toprule
        Case & $\overline{\dot{x}}_\text{A}$ & $\overline{\dot{y}}_\text{A}$ & $\overline{\dot{z}}_\text{A}$ & $\overline{\dot{\alpha}}$ & $\overline{\dot{\theta}}_1$ & $\overline{\dot{\theta}}_2$\\
        \midrule
        1 & 0.0307 & -0.0148 & 0.0182 & 0.0201 & 0.0240 & 0.0283\\
        2 & 0.0256 & 0.0009  & 0.0122 & 0.0036 & 0.0224 & -0.0917 \\
        3 & 0.0259 & -0.0020 & 0.0128 & 0.0024 & 0.0019 & -0.0724\\
        4 & 0.0253 & -0.0013 & 0.0127 & 0.0024 & 0.0047 & -0.0738\\
        \bottomrule
    \end{tabular}
    \caption{Average value of vehicle and manipulator velocities for each case. Linear velocities $\overline{\dot{x}}_\text{A}$, $\overline{\dot{y}}_\text{A}$, and $\overline{\dot{z}}_\text{A}$ are in m/s, and angular velocities $\overline{\dot{\alpha}}$, $\overline{\dot{\theta}}_1$, and $\overline{\dot{\theta}}_2$ are in rad/s.}
    \label{table:wln-velocity-average}
\end{table}

\begin{table}[ht]
    \centering
    \footnotesize
    \begin{tabular}{ c c c c }
        \toprule
        Case & $\overline{g_1}$ & $\overline{g_2}$ & $\overline{g_3}$ \\
        \midrule
        5 & \SI{0.9923}{} & \SI{2.4711e-4}{} & \SI{0.3053}{} \\
        6 & \SI{1.0000}{} & \SI{1.0017e-3}{} & \SI{0.2928}{} \\
        7 & \SI{0.9922}{} & \SI{8.9955e-5}{} & \SI{0.3039}{} \\
        8 & \SI{0.9926}{} & \SI{1.1569e-4}{} & \SI{0.2674}{} \\
        9 & \SI{1.0000}{} & \SI{9.9368e-5}{} & \SI{0.2661}{} \\
        \bottomrule
    \end{tabular}
    \caption{Average value of cost functions for each case.}
    \label{table:gpm-cost-average}
\end{table}
\section{Experimental Demonstration}
\label{section:experiments}

\subsection{Teleoperational Demonstration}
\begin{figure}[t]
    \centering
    \includegraphics[width=0.95\linewidth]{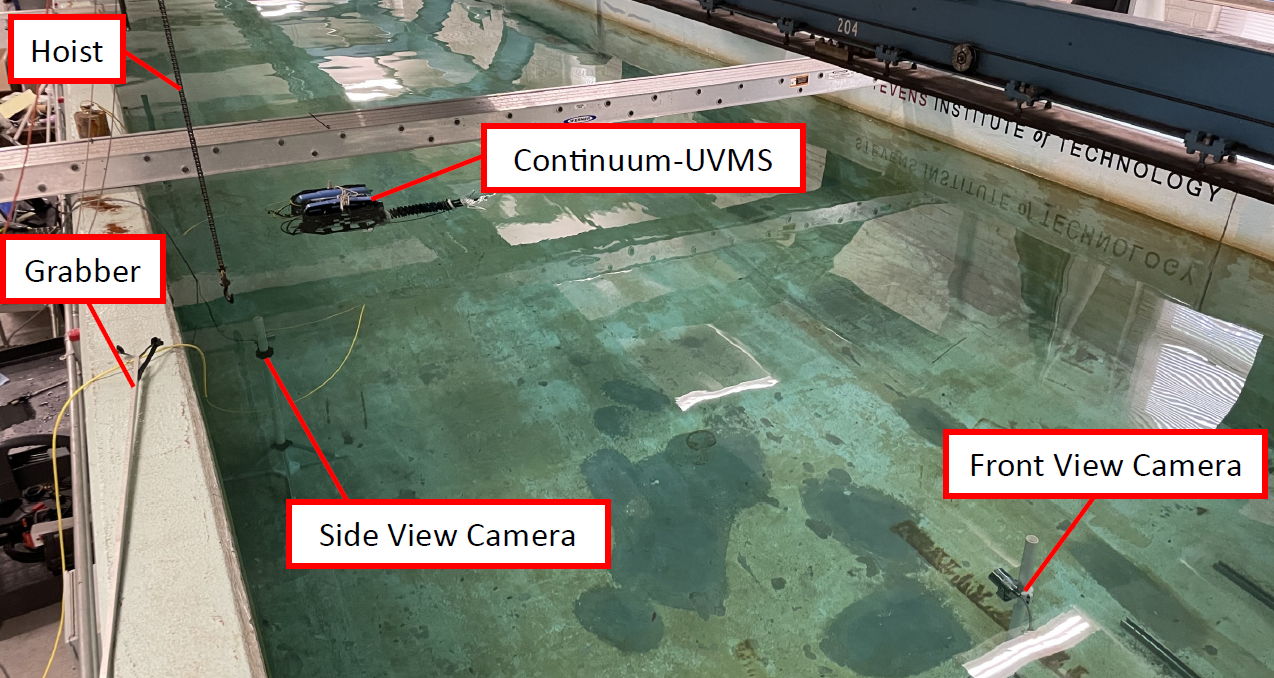}
    \caption{Experimental setup in Davidson Laboratory towing tank.}
    \label{fig:experiment-setup}
\end{figure}

The continuum-UVMS was deployed in the Davidson Laboratory towing tank at Stevens Institute of Technology. The experimental setup can be seen in Figure~\ref{fig:experiment-setup}. Initial testing of the UVMS was done using teleoperational control using a separate gamepad controller for each. This demonstrated three important capabilities. First, it showed the integrated continuum manipulator and vehicle were both waterproof. Second, it showed that communications between the manipulator, the vehicle, and the topside computer were successfully established. Third, it showed that arm movements had very little impact on the pose of the vehicle, which suggests that decoupling between the manipulator and the vehicle is a valid assumption.

\subsection{Autonomous Trajectory Demonstration}

A more sophisticated test of the continuum-UVMS is to autonomously execute the same trajectory simulated in Section~\ref{section:simulation}, where the robot moves from rest to grasp a stationary target at a known location. The coordinated vehicle and manipulator commands are generated in real time on separate devices, kept in-sync via ROS. The vehicle commands are generated by the topside computer and include the linear velocity, which is executed using an open-loop velocity controller, and the goal orientation, which is executed using a closed-loop position controller with IMU feedback. The manipulator commands generated by the onboard computer are executed using a separate closed-loop position controller in the actuation unit microprocessor using built-in servomotor encoders.  This experimental test provides a preliminary qualitative assessment of the feasibility of autonomous joint speeds distribution between the vehicle and the continuum manipulator.

\begin{figure}[t]
    \centering
        \includegraphics[width=0.95\columnwidth]{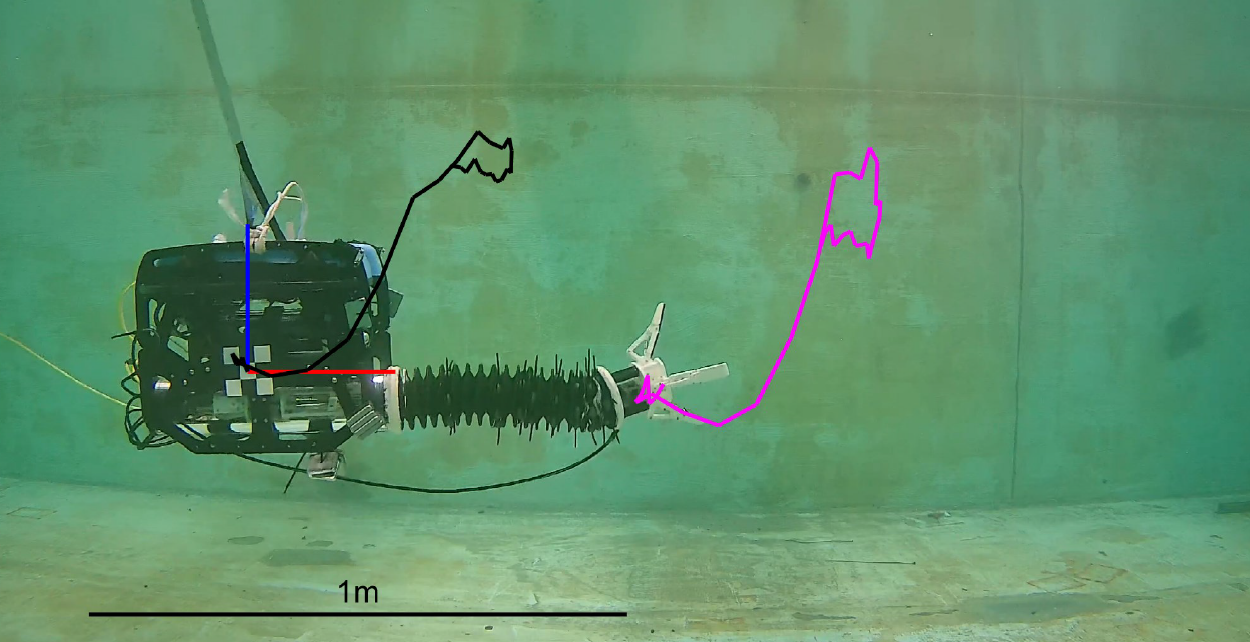}\\[6pt]
        \includegraphics[width=0.95\columnwidth]{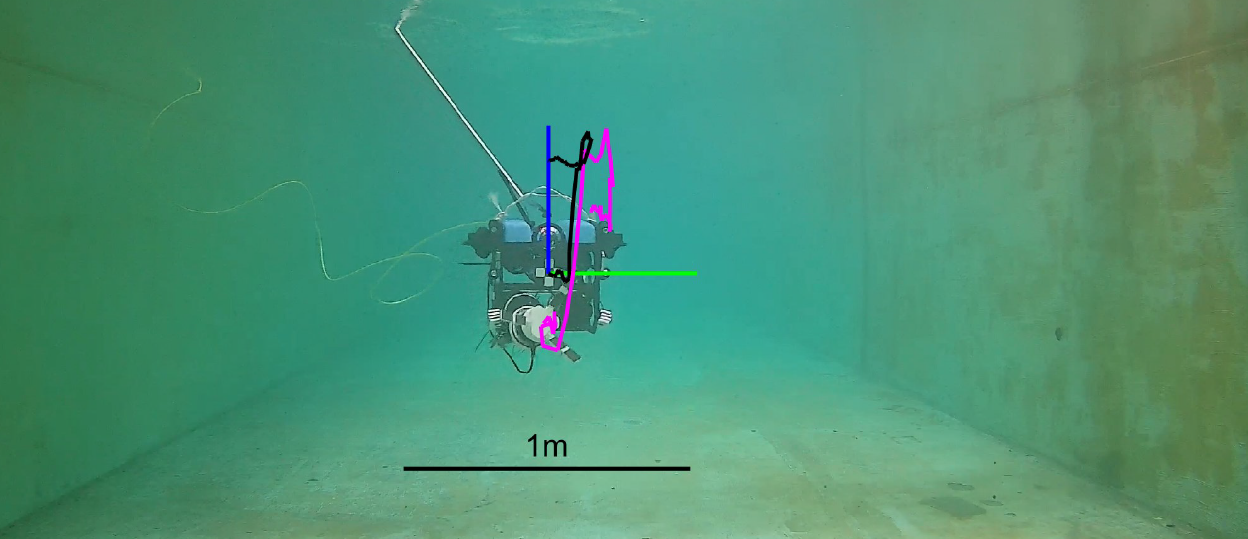}
    \caption{Continuum-UVMS in experimental deployment, with vehicle and end-effector trajectory marked in black and magenta, respectively.}
    \label{fig:experiment-trajectory}
\end{figure}

A time lapse of the end-effector trajectory can be seen in Figure~\ref{fig:experiment-trajectory}. The trajectory was captured using two underwater cameras, mounted perpendicularly to record both the global $xz$ and global $yz$ planes. The experimental trajectory of the arm is found by manually selecting the tip of the arm every 100 frames of the video playback. In Figure~\ref{fig:experiment-trajectory}, the vehicle trajectory is shown in black, and the end-effector trajectory is shown in magenta. The experimental trajectory showed overshoot when compared with the simulated trajectory, but followed the general desired motion.

\begin{figure}[!ht]
    \centering
    \captionsetup{justification=centering}
    \begin{subfigure}[b]{0.95\linewidth}
        \centering
        \includegraphics[width=\textwidth]{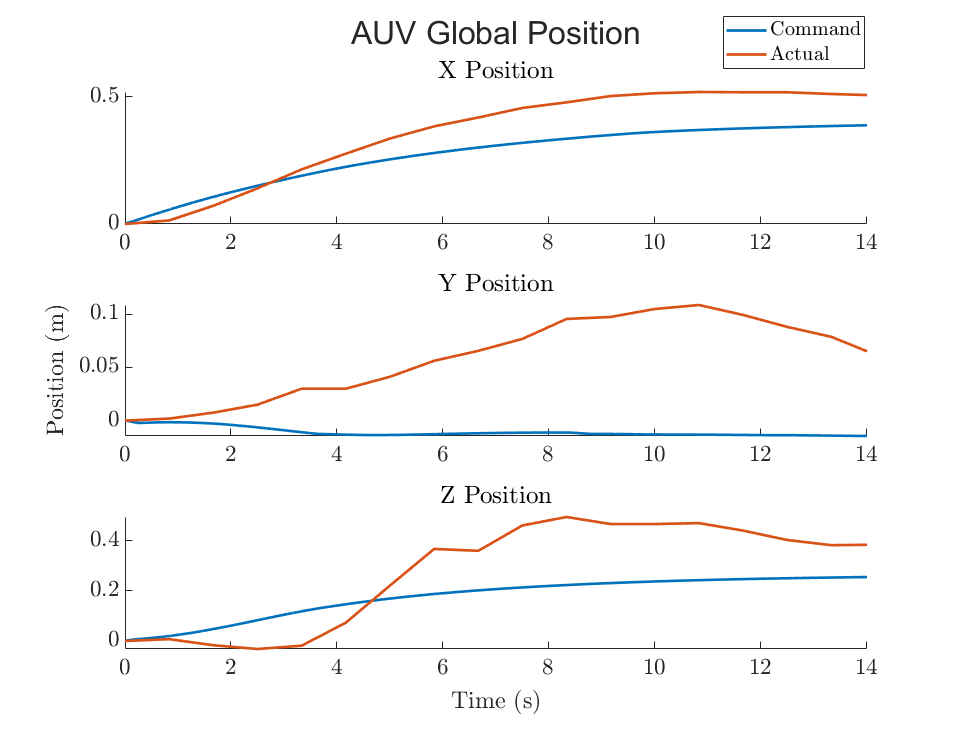}
        \caption{Vehicle command vs. actual position.}
        \label{fig:vehicle-error}
    \end{subfigure}
    \begin{subfigure}[b]{0.95\linewidth}
        \centering
        \includegraphics[width=\textwidth]{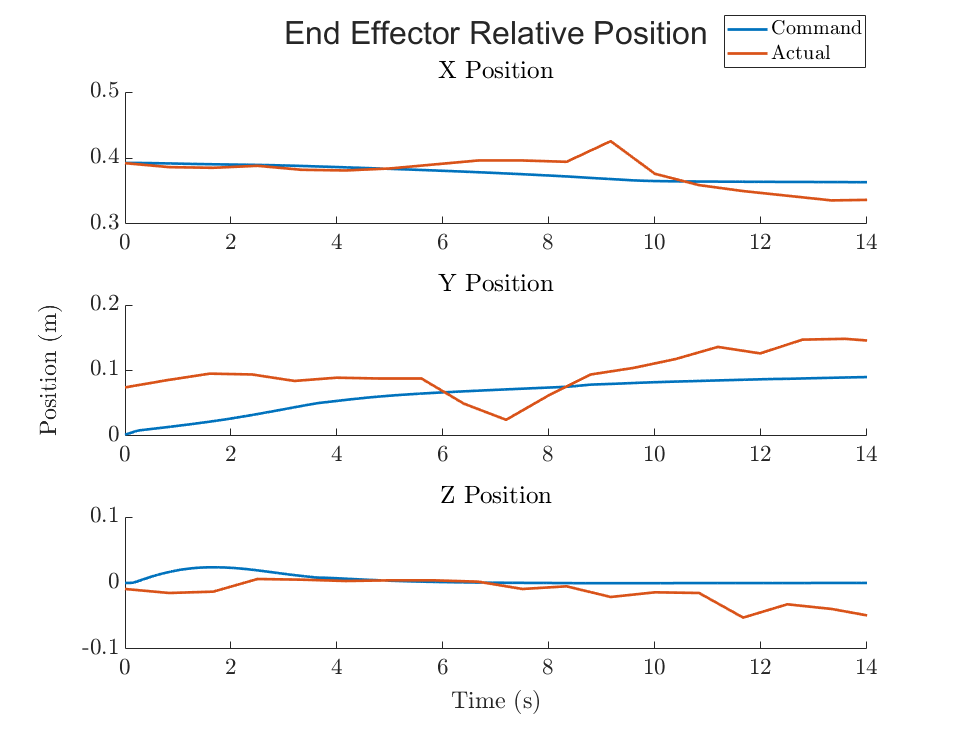}
        \caption{End-effector command vs. actual relative position.}
        \label{fig:end-effector-error}
    \end{subfigure}
    \caption{Experimental positioning error analysis.}
    \label{fig:experiment-error}
\end{figure}

The vehicle commanded position is plotted against the experimental trajectory in Figure \ref{fig:vehicle-error}. This plot indicates the vehicle drift which is not accounted for with the open-loop velocity control. In addition, overshoot in the Z direction of the vehicle is present which is likely due to the effects of buoyancy as the vehicle moves upwards during the trajectory and drift due to the inertia of the system. However, the positioning error of the manipulator relative to the base is more relevant to the scope of this paper and is shown in Figure \ref{fig:end-effector-error}. This plot shows a much higher manipulator accuracy between the end-effector position command and the experimental trajectory. Error in each of the principal axes is kept within \SI{0.05}{\m}. This is comparable with the end-effector positioning error reported in Ambar et. al. \cite{ambar2015experiment}, Cieslak et. al. \cite{cieslak2015autonomous}, and Heshmati-Alamdari et. al. \cite{heshmati2018robust}, whose reported error bounds are summarized in Table \ref{table:experiment-comparison}.

\begin{table}[ht]
    \centering
    \footnotesize
    \begin{tabular}{ c c }
        \toprule
        Publication & Error Bound \\
        \midrule
        Ambar 2015 & \SI{0.02}{\m} \\
        Cieslak 2015 & \SI{0.02}{\m} \\
        Heshmati-Alamdari 2018 & \SI{0.1}{\m} \\
        Our results & \SI{0.05}{\m} \\
        \bottomrule
    \end{tabular}
    \caption{End-effector experimental error comparison.}
    \label{table:experiment-comparison}
\end{table}

\subsection{Discussion}
There are several limitations to the proposed kinematic control algorithm that are evident in these demonstrations. These limitations can be broadly classified into two main categories: flaws in the control system hardware and unmodeled dynamics. The most significant flaw with the control system hardware is the lack of a navigation system. State-of-the-art systems that can provide real-time state estimation are quite expensive, and as a result we did not have access to such a system. The on-board IMU could in theory be used for dead reckoning, but this is quite noisy and prone to drift. Lacking adequate sensors or control for state feedback, we instead implemented UVMS control in our experiments using only open-loop control. The thruster gains were manually tuned to achieve the most accurate performance, but open-loop control fundamentally limits the accuracy of the experiments, and therefore future work would require development of a functional navigation system. Finally, the control of the vehicle is limited by a noticeable deadband in the velocity controller. This means that small, precise autonomous velocity commands to adjust the vehicle's position do not result in any output velocity, making accurate positioning control even more difficult.

The second shortcoming of the proposed method is the lack of a dynamic model for the UVMS, and thus lack of control for many potential sources of error. The magnitude of this impact is difficult to measure, but a few significant areas of potential error can be identified. First, the buoyancy of the UVMS in theory should be neutral; however, in practice, it is common to adjust the ballast so the vehicle is slightly positively buoyant in order to safely recover the vehicle in the event of loss of control. The vehicle depth controller compensates for this, but when the vehicle is given an upwards velocity command, the buoyancy tends to add extra lift which causes greater acceleration than desired. As mentioned previously, this is likely a significant source of error in the Z position overshoot of the vehicle in particular. The inertia of the vehicle is also a significant factor in positioning error which causes the vehicle to tend to overshoot the desired position and drift when given no commands. In addition, the inertial drift can cause the vehicle to not be fully at rest when initially executing the trajectory, which is a significant problem since we are using open-loop control. To mitigate the initial drift we used a long-handled grabber to stabilize the vehicle before executing the trajectory, which showed improvements in performance. Finally, the weight of the tether adds a small but noticeable backwards force on the vehicle, causing a slight drift backwards in the direction of the tether.

These limitations are rooted in the precise control and navigation of the vehicle specifically. Our contributions in this paper are the novel integration of a continuum manipulator with an underwater vehicle and the development of a kinematic model and controller for this continuum-UVMS. As a result, improvements to the vehicle positioning accuracy are outside of the scope of this paper, but are a very important area to focus on in future work.
\section{Conclusion}
\label{section:conclusion}

This paper introduces the novel continuum-UVMS, a free-floating underwater vehicle with a continuum manipulator. The mechanical and electrical design of the system is explained, as well as the integration of the continuum manipulator with the BlueROV2 platform. The kinematics and total Jacobian matrix are mathematically derived as part of the kinematic control algorithm, which generates optimal trajectories. Choices of the objective functions for subtask optimization are discussed and compared in simulation. The optimization of the trajectory is validated using a kinematic simulation. The experimental deployment and demonstration of the continuum-UVMS are then presented, with sources of error in the vehicle control discussed as well.

Future work should include improvements to the vehicle closed-loop controller to increase vehicle velocity accuracy and consequently reduce end-effector positioning error, which is vital in many real-world applications. Better tuning of algorithm parameters may allow the robot to better balance speed and performance, as well as improve optimization of secondary objectives. Investigation of new and more sophisticated tasks, and choice of related secondary objectives to optimize such tasks, is an important research area to explore the capabilities of this system. Multi-manipulator planning may also be a lucrative research topic, with potential applications including dexterous underwater manipulation, perching, or climbing.

\begin{acknowledgment}
This research was supported in part by USDA-NIFA Grant No. 2021-67022-35977.
\end{acknowledgment}

\bibliographystyle{IEEEtran}
\bibliography{formatting/sources}

\end{document}